\documentclass[12pt,a4paper]{article}

\usepackage{arxiv}

\usepackage[utf8]{inputenc} % allow utf-8 input
\usepackage[T1]{fontenc}    % use 8-bit T1 fonts
\usepackage{hyperref}       % hyperlinks
\usepackage{url}            % simple URL typesetting
\usepackage{booktabs}       % professional-quality tables
\usepackage{amsfonts}       % blackboard math symbols
\usepackage{nicefrac}       % compact symbols for 1/2, etc.
\usepackage{microtype}      % microtypography
\usepackage{lipsum}
\usepackage{graphicx}
\usepackage{float}
\usepackage{amsmath}

\title{Fourier Transform Approach to \\Machine Learning III: Fourier Classification \\}

\author{
  Soheil Mehrabkhani \\
\\
  \texttt{soheil.mehrabkhani@googlemail.com} \\}

\begin{document}
\maketitle

\begin{abstract}

We propose a Fourier-based learning algorithm for highly nonlinear multiclass classification. The algorithm is based on a smoothing technique to calculate the probability distribution of all classes. To obtain the probability distribution, the density distribution of each class is smoothed by a low-pass filter separately. The advantage of the Fourier representation is capturing the nonlinearities of the data distribution without defining any kernel function. Furthermore, contrary to the support vector machines, it makes a probabilistic explanation for the classification possible. Moreover, it can treat overlapped classes as well. Comparing to the logistic regression, it does not require feature engineering. In general, its computational performance is also very well for large data sets and in contrast to other algorithms, the typical overfitting problem does not happen at all. The capability of the algorithm is demonstrated for multiclass classification with overlapped classes and very high nonlinearity of the class distributions. 
 
\end{abstract}

% keywords can be removed

\vskip 0.5in

\section{Introduction}

For classification as one of the main areas of supervised learning, there are several successful algorithms developed. Some of the wide-spread algorithms for classification tasks are KNN [1-3], logistic regression [1-3], decision trees [1-3], random forest [1-3], naive Bayes classifier [1-3], Support vector machines [1-3] and Neural networks [4]. Depending on the feature dimensionality and the amount of the available data, the accuracy, the computational efficiency and the application areas of these algorithms are different. One of the challenging problems in most classification algorithms occurs if the algorithm must capture highly nonlinear decision boundaries. Of course, many of the classification algorithms can overcome such difficulties by methods like kernelization. However, the computational efficiency of such methods can be extremely low. Furthermore, the accuracy of such methods is usually highly dependent on several hyperparameters, which makes the development and the training of such algorithms rather cumbersome and time-consuming. Generally, to achieve the goal of artificial intelligence, not only the algorithms must come up with complex problems, which traditionally and exclusively belong to the area of human competence but it is necessary and desired to develop algorithms with minimal dependency on human intervention. Such algorithms would be not only intelligent in the sense of the handling tasks autonomously but they would be also intelligent in the way, how they are developed and trained. One candidate for approaching such autonomously trained algorithms is reducing the number of needed hyperparameter or at least applying less sensitive hyperparameter, which could be set to fixed values for a wide range of data sets.
Recently, we proposed Fourier-based algorithms for treating regression [5] and clustering [6] tasks, which are essentially smoothing approaches. In this work, we extend the application domain of Fourier-based learning to the classification tasks, especially for highly nonlinear decision boundaries. First, we introduce the formulation of the density function for multi-class classification and then we show, how the density function will be smoothed, and at the end how the probability distribution for all classes is related to the smoothed density function. 

\vskip 0.1in

\section{Fourier Classification Algorithm }
\label{sec:headings}
\ Consider the data set $C=\{\textbf{\textit{x}} ^i=(x_1^{(i)},x_2^{(i)}): i=1,2,...,N\}$ with $N$ distinct data points of all classes and two features $x_1$, $x_2$. We define the density function of the data set $C$ by a Dirac mixture Distribution [7] as follows:

\begin{equation}
\ \rho_k(x_1,x_2)=\sum_{\textbf{\textit{x}}^i\in C_k}^{N}\delta(x_1-x_1^{(i)},x_2-x_2^{(i)})-\sum_{\textbf{\textit{x}}^i \in C-C_k}^{N}\delta(x_1-x_1^{(i)},x_2-x_2^{(i)}),
\end{equation}

where $\delta(x_1-x_1^{(i)},x_2-x_2^{(i)})$ is the Dirac delta function [8] located at the point $(x_1^{(i)},x_2^{(i)})$ and $\rho_k(x_1,x_2)$ is the density function in respect to the the $k$-th class. We use an One-vs-rest [2] approach for multiclass claasification.  

Basically, the smoothness constraint may be imposed by a low pass filter [8], which can be the $\rho_k(x_1,x_2)$ to remove the noise, the spectrum of the $\rho_k(x_1,x_2)$ should be filtered by a Gaussian filter [9]. The $\tilde{\rho}(f_1,f_2)$ indicates the spectrum of the density function The $\tilde{\rho}(f_1,f_2)$ is the result of applying the Fourier transform (FT) on the $\rho_k(x_1,x_2)$:

\begin{equation}
\ \tilde{\rho}_k(f_1,f_2)=\mathcal{F}\{{\rho}_k(x_1,x_2)\}=\iint \limits_{x_1,x_2=-L/2}^{x_1,x_2=+L/2}dx_1dx_2\rho_k(x_1,x_2)e^{-2\pi j(x_1 f_1+x_2 f_2)},
\end{equation}

where $\mathcal{F}$ is the FT operator and $(f_1,f_2)$ is the frequency pair. The $L$ is the width of the spatial space enclosing the sampled data set $(x_1^{(i)},x_2^{(i)})$. Multiplying the spectrum of the density function $\tilde{\rho}_k(f_1,f_2)$ by a Gaussian distribution with the standard deviation $\tilde{\sigma}$ results in the spectrum of the smoothed density function (SDF):

\begin{equation}
\ \tilde{\rho}_{k,s}(f_1,f_2)=\tilde{\rho}_k(f_1,f_2)  \frac{1}{2\pi \tilde{\sigma}^2}e^{- (f_1^2+f_2^2)/2 \tilde{\sigma}^2}.
\end{equation}

The SDF${\rho}_s$ is calculated by applying inverse FT (IFT) on Eq. (4):

\begin{equation}
\ {\rho}_{k,s}(x_1,x_2)=\mathcal{F}^{-1}\{\tilde{\rho}_k(f_1,f_2)  \frac{1}{2\pi \tilde{\sigma}^2}e^{- (f_1^2+f_2^2)/2 \tilde{\sigma}^2}\},
\end{equation}

where $\mathcal{F}^{-1}$ is the IFT operator. Equation (5) can be converted to an equivalent equation in the spatial space by the use of the convolution theorem [10]:

\begin{equation}
\ {\rho}_{k,s}(x_1,x_2)={\rho_k}(x_1,x_2)*e^{-2\pi^2 \tilde{\sigma}^2(x_1^2+x_2^2)},
\end{equation}

where $*$ is the convolution operator. As can be seen in Eq. (6), the IFT of the Gaussian distribution is again a Gaussian function with a standard deviation $\sigma=1/(2\pi \tilde{\sigma})$. In principle, the convolution of the density function with the Gaussian function is responsible for the desired smoothness.
From plugging Eq. (1) in Eq. (6) and considering the definition of the Dirac Delta function follows:

\begin{equation}
\ {\rho}_{k,s}(x_1,x_2)=\sum_{i=1}^{N}\delta(x_1-x^{(i)},x_2-x_2^{(i)})*e^{-2\pi^2 \tilde{\sigma}^2(x_1^2+x_2^2)}=\sum_{i=1}^{N}e^{-2\pi^2 \tilde{\sigma}^2[(x_1-x_1^{(i)})^2+(x_2-x_2^{(i)})^2]},
\end{equation}

which shows, that basically, the SDF is a summation of $N$ Gaussian functions (with the standard deviation $\sigma$) placed at the $N$ data points $(x_1^{(i)},x_2^{(i)})$. However, Eq. (6) computationally is very inefficient, thus, in the algorithm, the ${\rho}_{k,s}(x_1,x_2)$ will be obtained by the use of FT and IFT in Eqs. (2) and (4), which are much more efficient: 

\begin{equation}
\ {\rho}_{k,s}(x_1,x_2)=\frac{1}{2\pi \tilde{\sigma}^2}\mathcal{F}^{-1}\{\mathcal{F}\{\sum_{i=1}^{N}\delta(x-x_1^{(i)},x_2-x_2^{(i))}\}  e^{- (f_1^2+f_2^2)/2 \tilde{\sigma}^2}\},
\end{equation}

Equation (8) is our estimate of the original density function corresponding to the total data. The standard deviation of $\tilde{\sigma}$ is the most significant unknown parameter, which still must be found by the algorithm.
Due to the applying, the discrete Fourier transform (DFT) [11] in the algorithm, the minimum spacing in the frequency space $df$ is $1/L$, thus, it is also used as the minimum value of the standard deviation in the frequency space. Consequently, the start value for the $\tilde{\sigma}$ is $1/L$  and its value in the $n$-th iteration is defined as follows:

\begin{equation}
\ \tilde{\sigma}^{[n]}=\frac{n}{L},
\end{equation}

where for normalized features, it is simply equal to the $n$ because $L=1$. Consequently, $\tilde{\sigma}$ in Eq. (7) can be replaced by the $n$:

\begin{equation}
\ {\rho}_{k,s}^{[n]}(x_1,x_2)=\frac{1}{2\pi n^2}\mathcal{F}^{-1}\{\mathcal{F}\{\sum_{i=1}^{N}\delta(x-x_1^{(i)},x_2-x_2^{(i))}\}  e^{- (f_1^2+f_2^2)/2 n^2}\}.
\end{equation}

Equation (9) shows the dependency of the SDF on the iteration number $n$. The higher $n_k$, the wider is Gaussian function in the frequency space, which allows considering higher frequencies, Thus, the higher iterations reduce the smoothness and simplicity of the function. In the language of machine learning it corresponds to the more complicated methods. For the first iteration $n=1$, the most higher frequencies are suppressed and the model is highly biased for higher iteration number the bias reduces. For too high iteration numbers, the standard deviation $\tilde{\sigma}$ of the Gaussian filter would be so high, that it wouldn't cut any frequencies at all and the SDF would be the same as the original density function. This would correspond to the overfitting problem.
The algorithm is expected to find the optimal number of iterations for SDF related to the $k$-th class, The optimal  $n$ for this class is indicated by the $n_k$. Clearly, for very low values of the $\tilde{\sigma}_n$, the number of significant frequencies contributing to the density function would be too low, thus the density function could not represent the significant variations in the sampled data, which are responsible for resolving the clusters. In other words, the corresponding model would be too simple and underfit the data. In contrast, overfitting the model could happen if it includes too high frequencies, which are basically responsible for extreme noise. To determine the best $\tilde{\sigma}_n$, the corresponding model will be evaluated by the correlation value of two successive smoothed density functions $corr(\rho_s^{[n]}),\rho_s^{[n-1]})$. If we consider its changes, the influence of the underfitting and overfitting may compensate for a $\tilde{\sigma}_n$. Thus, the algorithm stops where the second derivative of the correlation with respect to the iteration number reaches the minimum value of $\epsilon$:

\begin{equation}
\mid \frac{\partial^2 corr(\rho_s^{[n]},\rho_s^{[n-1]})}{\partial n^2}\mid< \epsilon.
\end{equation}

Inequality (9) is the convergence criterion for the algorithm and the $\epsilon$ is the convergence parameter, which practically is highly independent of the data sets and it can be fixed and for the most applications set to the value 0.01. Because the smoothing process must be repeated for all classes, the final $\tilde{\sigma}^{[n]}$ is defined as the maximum of the $\tilde{\sigma}^{[n]}$ values:

\begin{equation}
\ n_{final}=max\{n_k\} \quad k=1,2,...,K
\end{equation}

where $K$ is the number of classes. Now, the smoothed density functions for all classes are again computed by the use of Eq. (7) but with the fixed $\tilde{\sigma}_{final}$.
The FT and IFT in Eq. (7) will be accomplished by the use of a fast Fourier transform (FFT) [12], which is a fast algorithm for the implementation of the DFT. The standard DFT requires a uniform sampled data, however, in general, the given sampled data points $\{(x_1^{(i)},x_2^{(i)})\}$ are randomly distributed. Therefore, they must be mapped to an equidistant mesh. Consider the sets of both predictor values $\{x_1^{(i)}\}$ and $\{x_2^{(i)}\}$. Thus, we generate an equidistant quadratic mesh with $N_{mesh}$ points in both direction $x_1$ and $x_2$ and each data point will be mapped to the nearest mesh point. 

The mesh number must theoretically be sufficiently large to prevent deviations between the original predictor values and their new values caused by the mapping to the equidistant mesh. However, too large values of the spacing increase the number of the mesh points and consequently increase the computational complexity of the FFT. In principle, it is probable, that some data points have to be mapped to the same mesh point. The mesh points actually are the centers of the pixels with the area $dx_1dx_2$ and each pixel represents one single data point like a data point in the continuous space. Thus, for each mesh point, only one of the data points will be considered. It must be mentioned, that the Dirac delta function in the discrete form will be converted to the Kronecker delta function [11]. Consequently, the density value for pixels including a data point in one class will be one and for other points is zero.

In principle, the SDF ${\rho}_s(x,y)$ is expected to represent approximately the underlying probability distribution. Consequently, to satisfy the probability axioms, first, the minimum of all smoothed densities ${\rho}_{min}$ will be subtracted from the smoothed densities of all classes:

\begin{equation}
\ {\rho}_{k,s}(x_1,x_2)\xleftarrow{}{\rho}_{k,s}(x_1,x_2)-{\rho}_{min}.
\end{equation}

Now it will be normalized for all feature values, which results in the probability distributions of classes:

\begin{equation}
\ {P}_{k}(x_1,x_2)=\frac{{\rho}_{k,s}(x_1,x_2)}{\sum_{k=1}^{K}{\rho}_{k,s}(x_1,x_2)} ,
\end{equation}

\section{Results}
\ The proposed algorithm is applied to the data set presented in Fig. 1. Both predictors $x_1$ and $x_2$ are normalized. The data is composed of 3 classes and the noise is a Gaussian distribution with 3 different variances for each class. As can be seen in Fig. 1, the data distribution of classes is highly nonlinear and thus, nonlinear decision boundaries are required to separate classes. Furthermore, the non-convex form of distributions is an additional difficulty. The numbers of the used training data points for all classes are equal to prevent the data bias. 
The margins between the classes are very small and the classes are sightly overlapped. Thus, the classification of the data will be for the most classification algorithms very challenging or maybe impossible. 

\begin{figure}[H]
 \centering
 \includegraphics[width=10cm]{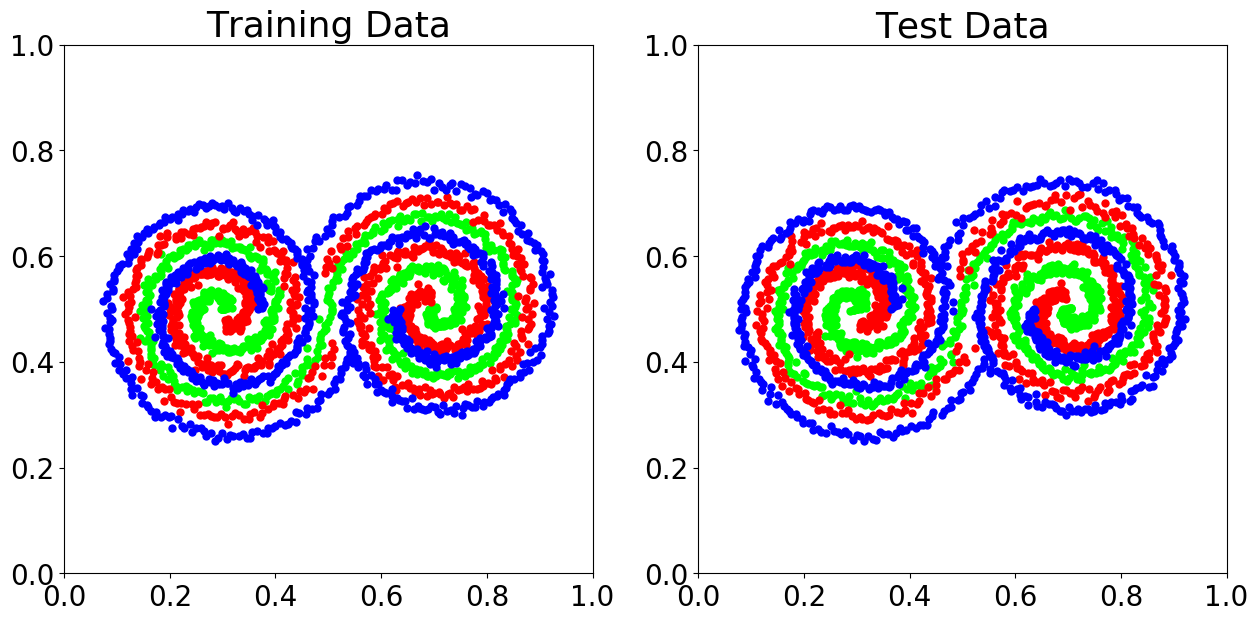}
 \caption{The data set is composed of 3 classes related to two normalized predictors $x_1$ and $x_2$.}
 \label{fig:soheil3}
\end{figure}

The equidistant mesh is generated with $N_{mesh}=512$ mesh points, which means a quadratic mesh with a total $512\times 512$ pixels. The first column in Fig. 2 shows the mapped data to the equidistant mesh for all 3 classes. According to Eq.(1) and considering the effect of the discretization on the Dirac function, the pixels value for the class $k$ is equal to $1$ and for all other classes is defined to be $-1$. The value for pixels belonging to no classes is set to zero. 
It should be noted that all these values after the smoothing process may be changed in each iteration because of applying different standard deviations $\tilde{\sigma}_n=n$. 

\begin{figure}[H]
 \centering
 \includegraphics[width=16cm]{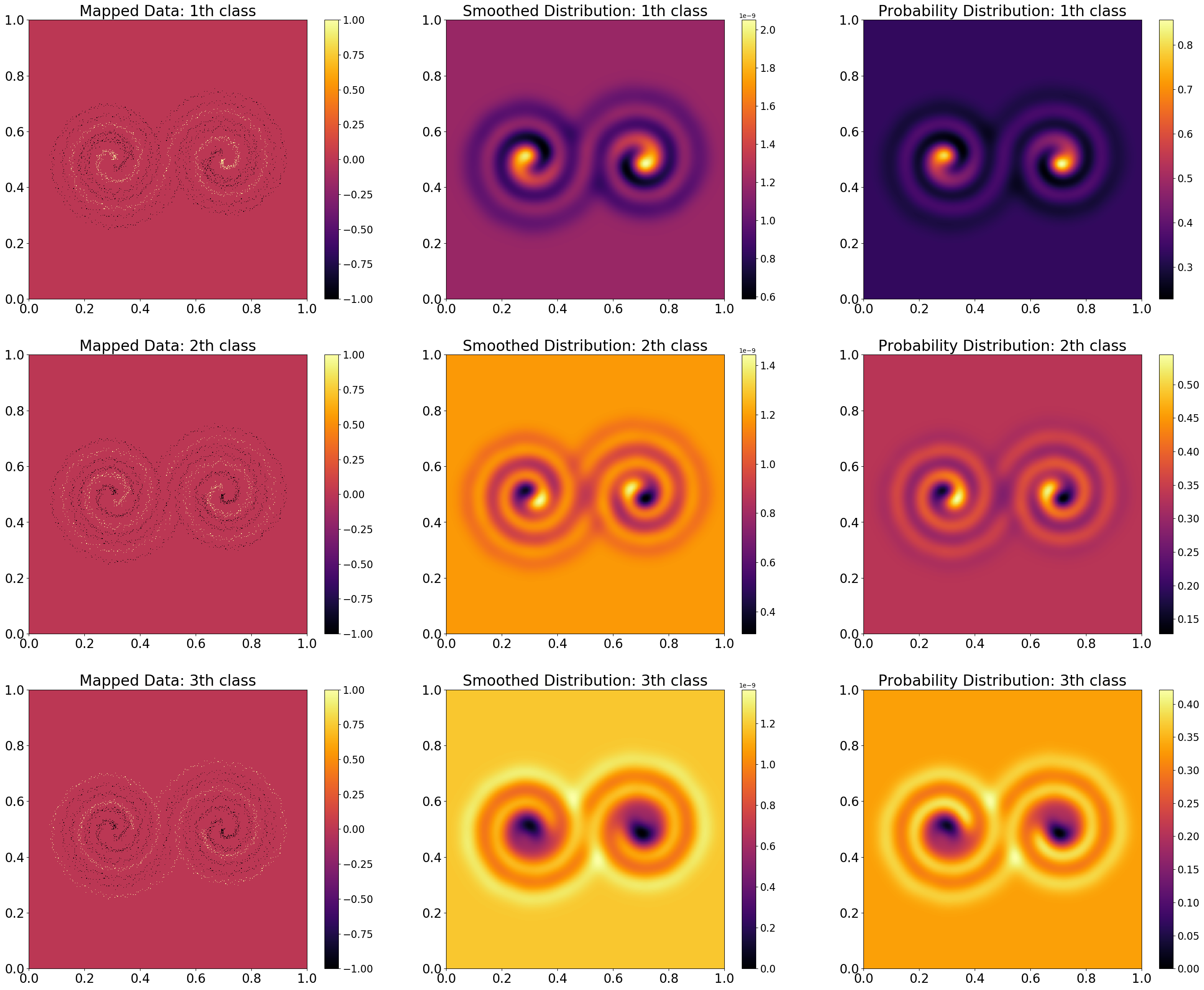}
 \caption{Probability Distribution of Classes}
 \label{fig:soheil1}
\end{figure}

The computed classifiers are shown in Fig. (3). 
The algorithm satisfied the convergence criterion (Inequality (10)) after only $4$ iterations with the convergence parameter $\epsilon=0.01$.

\begin{figure}[H]
 \centering
 \includegraphics[width=16cm]{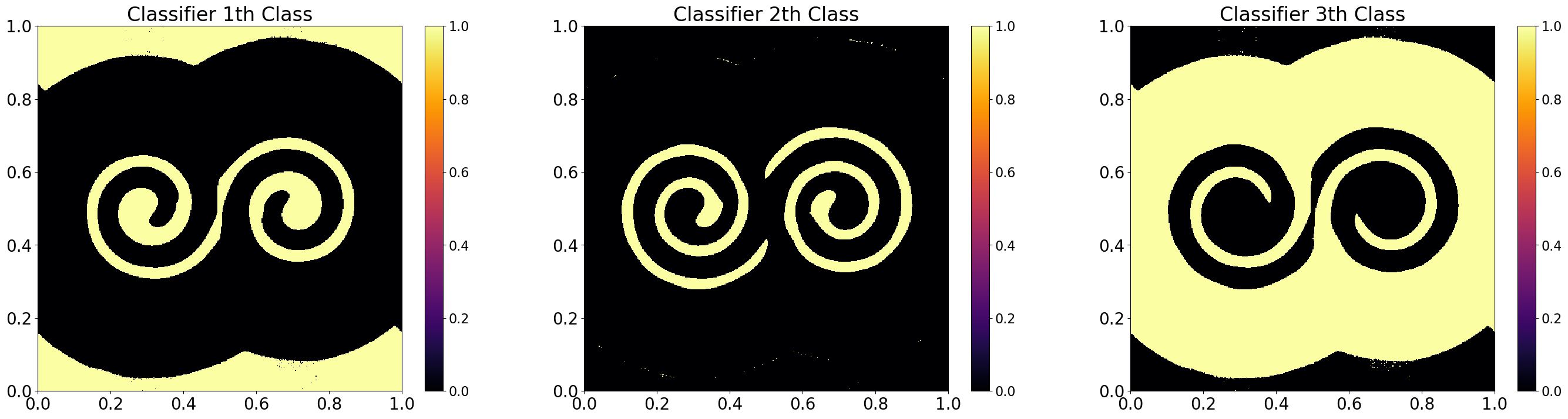}
 \caption{classifiers}
 \label{fig:soheil2}
\end{figure}

\section{Conclusion}

We presented a Fourier-based algorithm multiclass highly nonlinear classification tasks. The algorithm does not require any feature engineering or feature extraction prepossessing steps, which makes it very practical and time-consuming. The Fourier representation of the data does not assume any probability distribution, which makes it free of bias errors. Furthermore, the Fourier representation provides us with a complete description of the data, thus, the nonlinearity can be captured completely as long as the generated mesh is fine enough. The algorithm needs some hyperparameters, which in contrast to similar methods, have no significant influence on the overfitting or underfitting problems, but they can be set to fixed values without loss of generality. The mesh size can be changed if very high accuracies are desired. The usual underfitting and overfitting problems are excluded principally because of the nature of the algorithm.  The only drawback of the algorithm is its exponential dependency on the feature numbers, which makes it slow for very high dimensional classification tasks. However, because there is no requirement of data engineering for nonlinear modeling, only the original features must be considered not the higher powers of features or mixed terms. The applied metric for evaluation of the algorithm was recall with the values $97\%$ for the training data, and $96\%$ for the test data.


\begin{thebibliography}{1}

\bibitem{1}
T. Hastie and R. Tibshirani and
J. Friedman.
\newblock {\em The elements of statistical learning.} Springer press, 2th Edition 2008.

\bibitem{2}
C. M. Bishop.
\newblock {\em Pattern Recognition and
Machine Learning.} Springer, 2006.

\bibitem{3}
M. Sugiyama.
\newblock {\em Introduction to Statistical Machine Learning.} Elsevier, 2014.

\bibitem{4}
I. Goodfellow, Y. Bengio, A. Courville,
\newblock {\em  Deep Learning.} MIT Press, 2016.

\bibitem{5}
S. Mehrabkhani.
\newblock Fourier Transform Approach to Machine Learning I: Fourier Regression.
\newblock In {\em arXiv preprint arXiv:1904.00368}, 2019.

\bibitem{6}
S. Mehrabkhani.
\newblock Fourier Transform Approach to Machine Learning II: Fourier Clustering.
\newblock In {\em arXiv preprint arXiv:1904.13241}, 2019.

\bibitem{7}
O. C. Schrempf, D. Brunn, and U. D. Hanebeck.
\newblock Density Approximation Based on Dirac Mixtures with Regard to Nonlinear Estimation and
Filtering.
\newblock In {\em Proceedings of the 2006 IEEE Conference on Decision
and Control (CDC 2006)}, San Diego, California, Dec. 2006.

\bibitem{8}
D. W. Kammler
\newblock {\em A First Course in Fourier Analysis.} Cambridge University Press, 2007.

\bibitem{9}
E. Davies
\newblock {\em Computer and Machine Vision: Theory, Algorithms, Practicalities.} Academic Press, Elsevier, 2012.


\bibitem{10}
R. J. Marks II.
\newblock {\em The Joy of Fourier.} Baylor University, 2006

\bibitem{11}
I. Amidror.
\newblock {\em Mastering the Discrete Fourier Transform in One, Two or Several Dimensions.} Springer, 2013


\bibitem{12}
E. Brigham.
\newblock {\em Fast Fourier Transform and Its Applications.} Pearson, 1988



\end{thebibliography}
\end{document}